\documentclass[11pt]{article}

\usepackage[final]{acl}

\usepackage{times}
\usepackage{latexsym}
\usepackage{amsthm}
\usepackage{amssymb}
\usepackage{booktabs}
\usepackage{multirow}
\usepackage{algorithm}
\usepackage{algpseudocode}
\usepackage{tabularx}
\usepackage{amsmath}
\usepackage{enumitem}

\usepackage[T1]{fontenc}

\usepackage[utf8]{inputenc}

\usepackage{microtype}

\usepackage{inconsolata}

\usepackage{graphicx}

\title{AgenticRAG: Agentic Retrieval for Enterprise Knowledge Bases}

\author{Susheel Suresh\thanks{Equal contribution.}\thanks{sussuresh@microsoft.com}, Hazel Mak$^*$\thanks{hazelmak@microsoft.com}, Shangpo Chou, Fred Kroon, Sahil Bhatnagar \\ \\
        Microsoft Corporation }

\begin{document}
\maketitle

 \begin{abstract}
    We present \textbf{AgenticRAG}, a practical agentic harness for retrieval and analysis over enterprise knowledge bases. Standard RAG pipelines place significant burden of grounding on the search stack, constraining the language model to a fixed candidate set chosen deep in the retrieval process. Our approach reduces this overdependence by layering a lightweight harness on top of existing enterprise search infrastructure, equipping a reasoning LLM with search, find, open, and summarize tools enabling the model to iteratively retrieve information, navigate within documents, and analyze evidence autonomously. On three open benchmarks we observe substantial gains: 49.6\% recall@1 on BRIGHT (+21.8 pp over the best embedding baseline), 0.96 factuality on WixQA (+13\% relative improvement), and 92\% answer correctness on FinanceBench--within 2 pp of oracle access to true evidence. Ablation studies show that the most significant factor is the shift from single-shot retrieval to agentic tool use (5.9$\times$ improvement), while multi-query search and in-document navigation contribute to both quality and efficiency. We present various design choices in our agentic harness that were informed by pre-production deployments. Our results demonstrate its suitability for real-world enterprise production environments.

\end{abstract}

\section{Introduction}
\label{sec:intro}
    Standard retrieval-augmented generation (RAG) pipelines follow a static retrieve-then-generate paradigm \cite{lewis2020rag}. In this design, the search stack effectively determines the final candidate set the large language model (LLM) will see and the model’s reasoning is constrained to that set. Modern enterprise-grade search stacks are highly optimized for scalability, latency, and multi-stage ranking pipelines built on inverted indexes, probabilistic retrieval, and learned ranking models \cite{liu2009learning, nogueira2019passage, thakur2021beir}. These systems excel at keyword and short semantic queries and are strong for high-recall candidate generation. However, they are not designed to resolve situational, multi-document, or analytically complex information needs—the kinds of queries knowledge workers issue against dense corpora such as technical manuals, compliance documents, and financial reports. 

Real-world RAG systems \cite{microsoftAgenticRetrieval} attempt to compensate for these limitations through retrieval enhancement techniques such as HyDE \cite{gao2023precise}, multi-query reformulation \cite{wang2023query2doc}, and adaptive or iterative retrieval strategies \cite{trivedi2023interleaving, jeong2024adaptive}. While these methods provide robustness to query phrasing and higher retrieval coverage, they largely preserve the same architectural assumption: retrieval decisions are finalized before substantive reasoning begins. The LLM still operates over a fixed candidate set selected deep in the search stack, without the ability to iteratively navigate documents, synthesize evidence across sources, or reassess results from a higher-level vantage point.

Recent advances in reasoning-capable language models have demonstrated strong performance on planning and iterative external tool use \cite{yao2023react, schick2023toolformer}. Rather than hard-coding retrieval steps, we can empower the model itself to drive the process---deciding what to search for, which documents warrant deeper investigation, and when sufficient evidence has been gathered. This relaxes the pressure on the search stack: it only needs to achieve good recall, while the model handles the final precision from its broader context. We present \textit{AgenticRAG}, a practical harness that equips a reasoning LLM with four tools---\textbf{search}, \textbf{find}, \textbf{open}, and \textbf{summarize}---layered on top of existing enterprise search infrastructure. The search tool delegates to the underlying search stack for broad candidate discovery, while find and open serve as precision instruments that let the model drill into candidate documents via in-document search and full-content retrieval (with rolling window access). To manage the growing context during long reasoning chains, the harness monitors token usage and triggers the summarize tool when a threshold is reached, allowing the model to consolidate its findings while preserving key references. Our contribution is system-level: a lightweight inference-time tool harness that requires no model fine-tuning, custom embedding model, graph construction, or corpus-specific preprocessing beyond indexing documents into the existing enterprise search backend.

We evaluate on three benchmarks spanning retrieval, enterprise QA, and financial document reasoning. Our approach achieves 49.6\% recall@1 on BRIGHT (+21.8 pp over the best embedding baseline), 0.96 factuality on WixQA (+13\% relative), and 92.00\% answer correctness on FinanceBench--within 2 pp of oracle access. Our method is deployed for pre-production evaluation, and learnings from these deployments directly inform our design choices. We provide detailed ablations analyzing the contribution of each tool, the effect of multi-query search, and model-level differences in retrieval strategy.

\section{Related Work}
\label{sec:related}
Retrieval-Augmented Generation (RAG) grounds LLM generation in external corpora to mitigate parametric memory limitations \cite{lewis2020rag, guu2020realm}. Early approaches focused on identifying relevant documents using sparse or dense vector retrieval \cite{khattab2020colbert, izacard2021leveraging} to enhance performance on knowledge-intensive NLP tasks. As context windows expanded, research shifted toward scaling retrieval to trillions of tokens \cite{borgeaud2022improving} and optimizing in-context learning \cite{ram2023incontext, shi2023replug}. Despite these advancements, standard RAG pipelines often struggle with "long-tail" knowledge and can suffer from hallucinations when retrieval fails \cite{mallen2023when, gao2024retrieval}. Furthermore, static "retrieve-then-generate" paradigms lack the flexibility to handle complex, multi-hop queries that require iterative information gathering \cite{jiang2023active, press2023measuring}.

To address the brittleness of static pipelines, the field has evolved toward \textit{Agentic} patterns, where autonomous agents (LLMs) dynamically orchestrate the retrieval process \cite{singh2025agentic, oche2025systematic}. Foundational work in agentic behaviors, such as ReAct \cite{yao2023react} and Toolformer \cite{schick2023toolformer}, demonstrated that LLMs could effectively wield external tools to solve reasoning problems. This paradigm has been formalized in systems like Self-RAG \cite{asai2024selfrag} and Corrective RAG \cite{yan2024corrective}, which employ self-reflection mechanisms to critique retrieved content and trigger fallbacks (e.g., web search) when necessary. Recent approaches propose to integrate retrieval into planning: PlanRAG \cite{lee2024planrag} and Search-o1 \cite{li2025searcho1} separate high-level planning from low-level execution, allowing agents to decompose complex queries into sub-tasks. Similarly, Search-R1 \cite{jin2025searchr1} uses reinforcement learning to train LLMs for autonomous search decisions. While effective, many of these systems are designed for open-domain search or require fine-tuning, reinforcement learning, or dedicated retrieval policies, which makes them less directly applicable to proprietary enterprise corpora that cannot be exported for training. They can also incur high latency and token costs due to recursive reasoning loops \cite{trivedi2023interleaving}.

Another critical limitation in standard RAG is the "flattening" of documents into disjointed chunks, which discards valuable structural priors like headings and document boundaries. RAPTOR \cite{sarthi2024raptor} addresses this by recursively clustering and summarizing text chunks into a tree structure, enabling retrieval at varying levels of abstraction. Similarly, HiQA \cite{chen2024hiqa} constructs multi-document hierarchical contexts. Graph RAG \cite{edge2024local, scaffidi2025graphrag} approaches seek to build knowledge graphs from documents to support query-focused summarization. While powerful for unifying knowledge \cite{pan2024unifying, wang2024knowledge}, graph construction is often computationally prohibitive for dynamic enterprise environments.
In contrast, our Agentic RAG harness is an inference-time system that leverages a reasoning model with a "search" tool (using a fast enterprise grade search stack) alongside "find" and "open" tools for deeper information gathering and reasoning. This positions the contribution as a deployable system integration for enterprise file systems: it works with existing search infrastructure, preserves document access controls, and avoids extensive pre-computation or retraining.


\section{Method}
\label{sec:method}
    \subsection{System Overview}
\label{sec:system_overview}

We present an agentic RAG system for enterprise document search and question answering over large file systems. Unlike traditional single-pass RAG pipelines, our system employs an iterative reasoning loop where a large language model (LLM) autonomously decides when to search for documents, drill into specific passages, and retrieve full content before producing a final answer.

The system addresses several challenges in enterprise RAG: (1) multi-step reasoning: complex queries require information from multiple documents, (2) context window constraints: accumulated retrieval results must fit within LLM limits, (3) grounded responses: answers must include traceable citations to source documents, and (4) multi-turn efficiency: follow-up queries should reuse previously retrieved content rather than re-executing redundant searches. Our architecture supports multiple model families and reuses existing search infrastructure for the backend implementation of the retrieval tools. By lightweight, we mean that the harness consists of four tools, requires no model fine-tuning, no graph construction, and no custom embedding index beyond the enterprise search stack already deployed for document discovery. Overall the system comprises three main components:
\begin{enumerate}
  \item \textbf{Agentic Loop}: Orchestrates LLM-tool interactions, bounded by maximum iterations.
  \item \textbf{Retrieval Tools}: Three tools (\textit{search}, \textit{find}, \textit{open}) provide hierarchical access to enterprise documents. A \textit{summarize} tool for context management during long reasoning chains.
  \item \textbf{Conversation State}: Maintains message history, token accounting, and reference ID mappings that track documents across iterations. 
\end{enumerate}

\subsection{Agentic Loop}
\label{sec:agentic_loop}

\begin{figure}[t]
\centering
\includegraphics[width=\columnwidth]{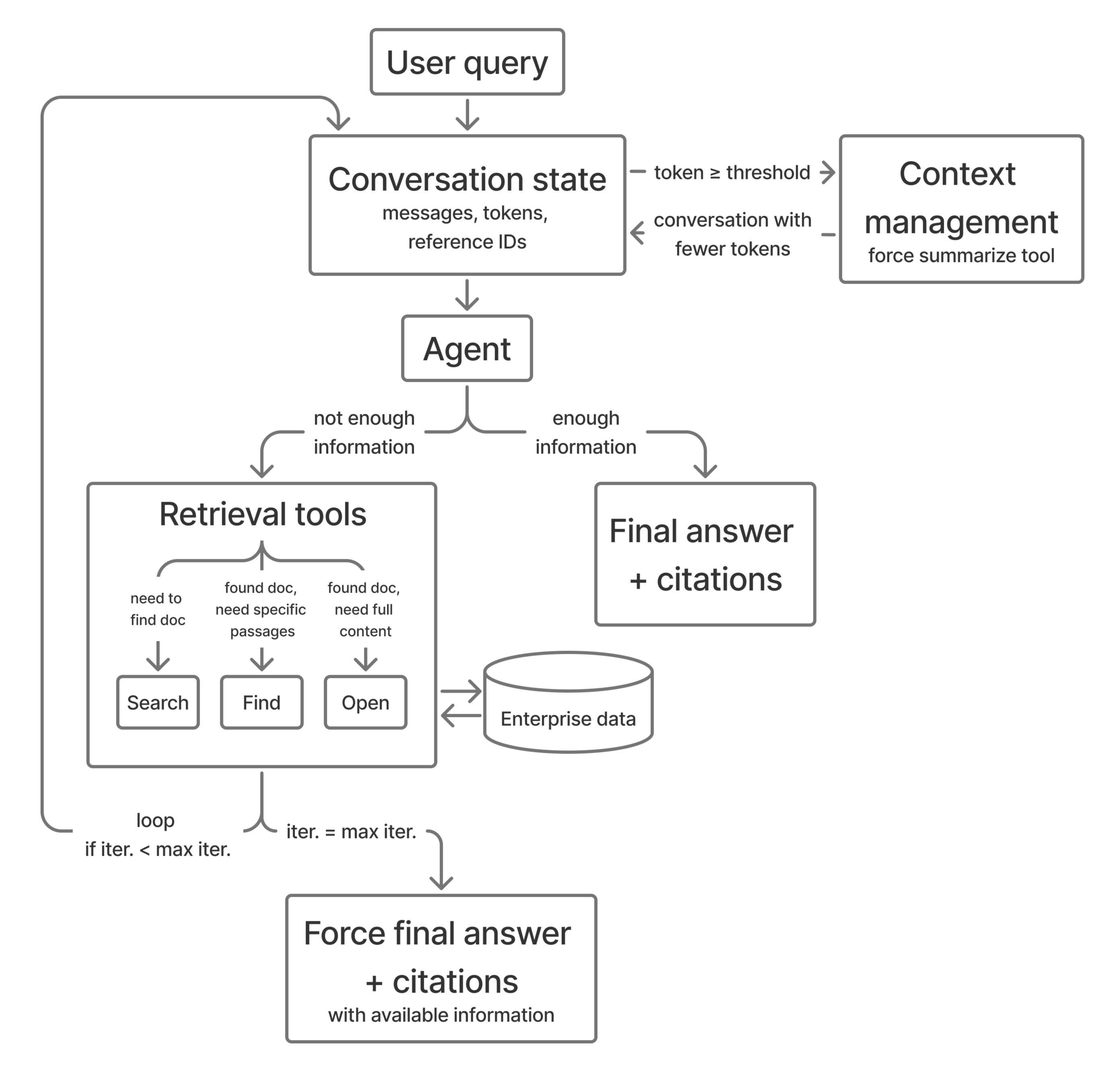}
\caption{Agentic loop}
\label{fig:agentic_loop}
\end{figure}

The agent processes each query through a bounded iteration loop (Figure \ref{fig:agentic_loop}). Each iteration, upon receiving the current conversation, the agent either selects a tool to call and appends to the conversation, or returns the final answer with citations.

The loop terminates under two conditions: (1) the model produces a text response, or (2) the iteration count reaches maximum iterations (default: 15). When maximum iterations are reached without a final answer, the agent issues a forced completion request, requiring the model to respond using available information. If the token budget is exceeded during execution, the agent triggers context management (Sec. \ref{sec:context_management}) to free space and continues the loop. For detailed algorithm, see Appendix \ref{app:agentic_loop_algo}.

\subsection{Retrieval Tools}
\label{sec:retrieval_tools}

The system provides three retrieval tools enabling hierarchical document exploration (Table \ref{tab:tool_def}). The agent decides which to invoke based on current information needs.

\textbf{Search} performs enterprise-wide document discovery by delegating to the existing enterprise search stack. In the default configuration, the model may issue up to five query reformulations in one tool call. The tool returns up to 10 results per query, each containing a snippet, title, filename, file type, and other available metadata. Results from multiple queries are combined and deduplicated. Each result receives a unique reference ID (format: turn$m$search$n$) using a globally incrementing counter, enabling subsequent find and open operations. 

\textbf{Find} performs targeted in-document search within a single document identified by its reference ID. Given a list of keyword patterns, lexical matching is case-insensitive substring matching; an optional semantic find mode can also be enabled. The tool returns up to 2 matching passages per pattern. Results are deduplicated by content and truncated at a bounded token limit ($\sim$11k tokens). Find is most useful when the model knows \emph{what} to look for, such as a revenue metric or a named concept inside a long filing.

\textbf{Open} retrieves full document content in a fixed line window. Each call returns a window of lines (default: 1,800) starting from either the beginning (line 0) or a specific line number chosen by the agent, and a response header indicating the viewing range and total document length (e.g., "Viewing lines [0--1799] of 3000 lines"). To access more than one portion from a file, the model makes subsequent calls with an explicit line number value. This enables navigation through documents exceeding the window size while keeping each response bounded. Open is most useful when the model knows \emph{where} to read, such as context around a table, section heading, or line-numbered preview. The system prompt guides effective tool usage. See Appendix \ref{app:system_prompt} for details. 

\begin{table*}[t]
\centering
\small
\caption{Retrieval Tool Specifications}
\label{tab:tool_def}
\small
\begin{tabularx}{\textwidth}{@{}lXXX@{}}
\toprule
\textbf{Tool} & \textbf{Definition} & \textbf{Input} & \textbf{Output} \\
\midrule
\textsc{search} & Discover relevant documents from entire corpus & queries & Snippets with reference ID and file metadata ($\leq$10 per query) \\
\addlinespace
\textsc{find} & Locate specific information from single document & reference id, patterns & Passages ($\leq$2 per pattern, $\leq$11k tokens total) \\
\addlinespace
\textsc{open} & Retrieve windowed full content from single document & reference id, line number (optional) & Line-numbered content ($\leq$1800 lines) \\
\bottomrule
\end{tabularx}
\end{table*}

\subsection{Context Management}
\label{sec:context_management}

\begin{figure}[h]
  \centering
  \includegraphics[width=\linewidth]{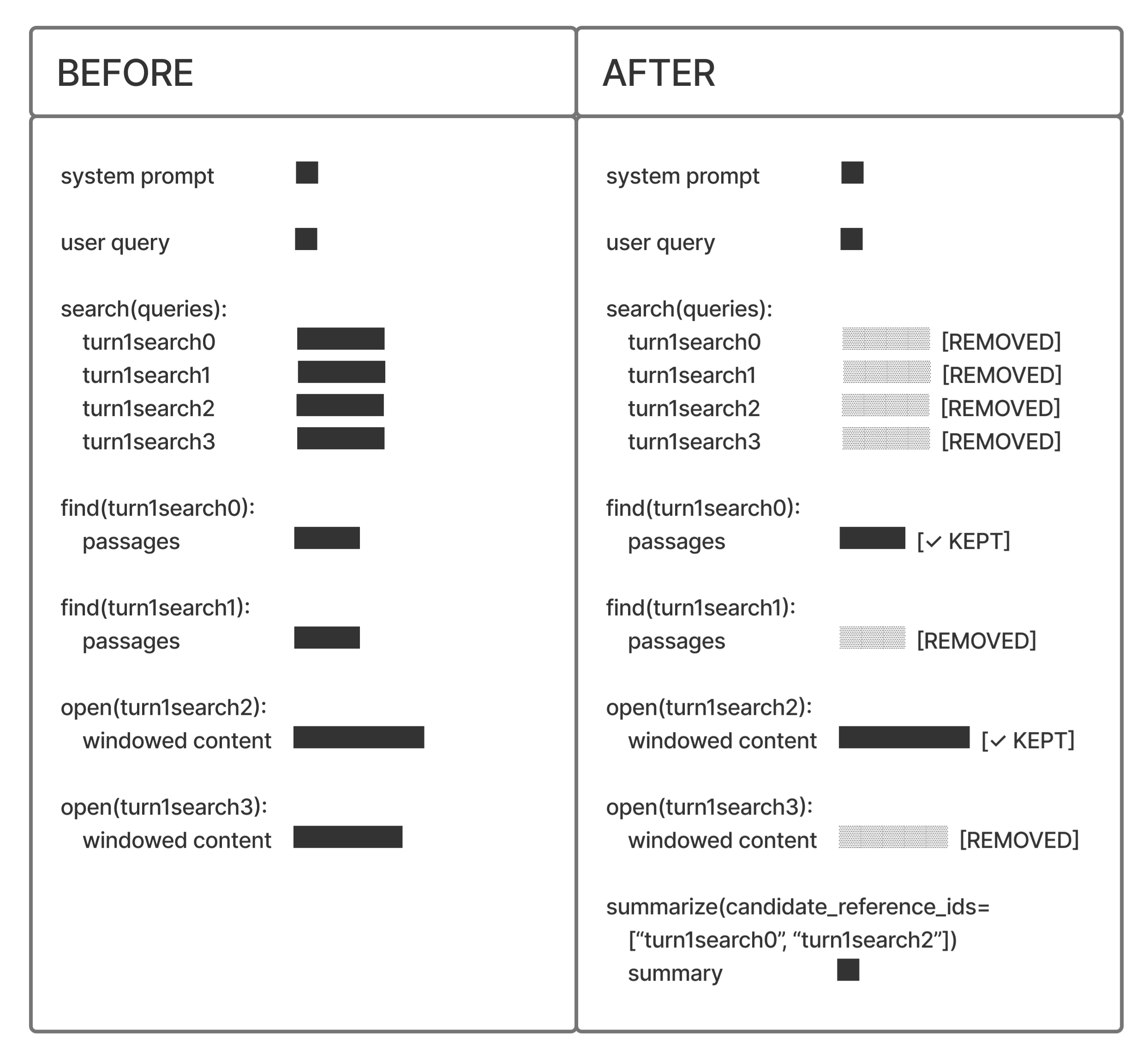}
  \caption{Example conversation history with context management via forced summarize tool call.}
  \label{fig:context_management}
\end{figure}

Since retrieval tools can load $\sim$11k tokens from files each time, the context window can be used up quickly. To manage that, the harness monitors token usage against a 128K-token threshold: it emits an internal warning when the conversation reaches 90\% of the budget and forces summarization at the threshold. The summarize tool enables the model to record current reasoning and designate which references to preserve. The system then scans tool messages and removes content not associated with preserved reference IDs, freeing tokens while retaining cited evidence. This approach extends effective context capacity. See Figure \ref{fig:context_management} for an example conversation history before and after context management.

\section{Experiment Setup}
\label{sec:exp_setup}
    Our goal is to evaluate AgenticRAG in realistic enterprise settings where knowledge workers issue complex situational queries requiring multi-step reasoning over large corpora of long, domain-specific documents. To this end we adopt the \textbf{BRIGHT benchmark} \cite{su2024bright} which contains StackExchange questions spanning eight domains. We choose to evaluate on the long-context setting of BRIGHT, where documents correspond to entire web pages rather than snippets and the task is to retrieve the full relevant document(s) for a given query. For our agentic setting, the full BRIGHT web pages are converted to document files and indexed into the same enterprise search backend used by the search tool. Search returns snippet previews with metadata and reference IDs, and the find and open tools then access full document content through those IDs. Standard protocol of using Recall@1 to measure relevance of retrieved documents is adopted \cite{su2024bright}. We instruct the model in AgenticRAG to provide relevancy scores for the citations it uses when producing the answer, which induces a ranking over documents for evaluation. We also test our method on \textbf{WixQA} \cite{cohen2025wixqa}, which targets real-world support and troubleshooting enterprise scenarios that require multi-document and multi-step reasoning for procedural answers. We adopt the same LLM based factuality metric defined in WixQA. Finally, we run evaluations on the popular \textbf{FinanceBench} \cite{islam2023financebench} dataset, which contains financial questions that require deep reasoning over large company financial documents. Our metric here is answer correctness as a proxy for accurate information retrieval, since questions pertain to single documents. Detailed benchmark descriptions, query set and corpus statistics are presented in Appendix \ref{app:datasets}.

\section{Results}
\label{sec:results}
    \subsection{Long-Context Retrieval on BRIGHT}
\label{sec:results_bright}

\begin{table*}[t]
\centering
\small
\caption{Long-context retrieval performance on unsplit web pages of StackExchange data from BRIGHT benchmark. Scores are reported in recall@1. Best baseline per category shown; full results in Table~\ref{tab:bright_model_compare_full}.}
\label{tab:bright_model_compare_main}
\begin{tabular}{llccccccccc}
\toprule
Category & Model & Bio. & Earth. & Econ. & Psy. & Rob. & Stack. & Sus. & Pony & Avg. \\
\midrule
Sparse & BM25 & 10.7 & 15.4 & 10.7 & 8.4 & 7.4 & 22.2 & 10.7 & 5.4 & 11.4 \\
Open-source Emb. & Qwen & 39.2 & 36.1 & 25.7 & 42.3 & 21.3 & 23.5 & 33.1 & 1.3 & 27.8 \\
Proprietary Emb. & Voyage & 34.4 & 35.4 & 26.7 & 41.6 & 12.9 & 12.8 & 31.1 & 1.3 & 24.5 \\
Reasoning Enhanced & ReDI & 28.4 & 22.4 & 21.2 & 32.0 & 19.8 & 36.3 & 21.7 & -- & 26.0 \\
\midrule
\multirow{2}{*}{\shortstack[l]{Ours (AgenticRAG)\\$\llcorner$search, find, open, summ.}} & GPT-5-mini & 61.7 & 48.1 & 41.4 & 65.3 & 39.4 & \textbf{40.6} & 46.6 & 4.8 & 43.5 \\
 & \textbf{Claude Sonnet 4.5} & \textbf{62.3} & \textbf{60.0} & \textbf{58.7} & \textbf{67.9} & \textbf{55.0} & 34.1 & \textbf{51.7} & \textbf{7.1} & \textbf{49.6} \\
\bottomrule
\end{tabular}
\end{table*}

Table~\ref{tab:bright_model_compare_main} presents retrieval performance on the BRIGHT benchmark, showing the best baseline per category. Full results with all models are in Appendix Table~\ref{tab:bright_model_compare_full}, including reasoning-enhanced baselines such as LLM re-rankers over BM25/SBERT and ReDI. Our agentic harness equipped with search, find, open, and summarize tools enables both Claude Sonnet 4.5 and GPT-5-mini to achieve the highest recall@1 across all eight benchmark splits compared to all baselines. With our AgenticRAG retrieval harness, Claude Sonnet 4.5 achieves \textbf{49.6\%} average recall@1 (+21.8 pp over Qwen, the best embedding model at 27.8\%) and GPT-5-mini reaches 43.5\% (+15.7 pp). Gains are consistent across domains, with the largest improvements in Economics (+33.0 pp), Earth Science (+24.6 pp), Robotics (+33.7 pp), and Psychology (+25.6 pp). Even with a vast corpus of 5.65K long documents averaging 16K tokens each, our method scales by leveraging traditional retrieval via the search tool, deeper reasoning enabled by the open / find tools and effective context window management by the summarize tool. The best-performing reasoning-enhanced baseline, ReDI, uses a fine-tuned Qwen3-8B decomposition and retrieval-fusion model and achieves 26.0\% recall@1 in the BRIGHT long-document setting. Our method outperforms it by +23.6 pp (via Claude) and +17.5 pp (via GPT-5-mini). Static approaches such as one-time query rewriting or LLM-based re-ranking cannot match the iterative reasoning that our harness provides, and the gap is evident across all splits.

\subsection{Enterprise QA on WixQA}
\label{sec:results_wixqa}
\begin{figure}[t]
\centering
\includegraphics[width=0.7\columnwidth]{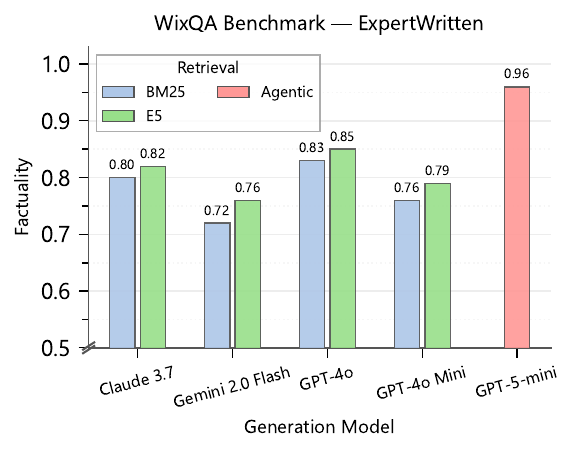}
\caption{Factuality performance on the WixQA Expert Written dataset (N=200). Our agentic approach (red) substantially outperforms BM25 (blue) and E5 (green) retrieval baselines across all generation models.}
\label{fig:wixqa_expertwritten}
\end{figure}

Figure~\ref{fig:wixqa_expertwritten} presents factuality results on the WixQA benchmark, which requires multi-document analysis to answer enterprise support questions. Semantic embeddings alone fail to capture the cross-document reasoning needed for these queries, whereas the iterative search and reasoning enabled by our harness excels. On the \textit{Expert Written} split, our method with GPT-5-mini achieves a factuality score of \textbf{0.96}, compared to 0.85 for E5 retrieval and 0.83 for BM25---a \textbf{13\%} relative improvement over the best baseline. We observe similar gains on the \textit{Simulated} split; see Appendix~\ref{app:wixqa_simulated_results} for details.

\subsection{Financial Document QA on FinanceBench}
\label{sec:results_financebench}

\begin{table}[htbp]
\centering
\small
\caption{Evaluation results on FinanceBench (N=150).}
\label{tab:financebench_results}
\begin{tabular}{lc}
\toprule
Method & Ans. Correct (\%) \\
\midrule
Traditional RAG & 24.24 \\
\midrule
\shortstack[l]{Agentic w. keyword search tools \\$\llcorner$pdfgrep,rga,linux cmd} & 32.71 \\
\midrule
\shortstack[l]{Golden Evidence + GPT-5-mini\\$\llcorner$oracle retrieval (full page)} & 94.00 \\
\midrule
\shortstack[l]{Ours (AgenticRAG) GPT-5-mini\\$\llcorner$search, find, open, summ.} & \textbf{92.00} \\
\midrule
\shortstack[l]{Ours (AgenticRAG) Claude Sonnet 4.5\\$\llcorner$search, find, open, summ.} & \textbf{91.78} \\
\bottomrule
\end{tabular}
\end{table}

Table~\ref{tab:financebench_results} presents answer correctness on the FinanceBench dataset, which evaluates question answering over real-world financial filings. The retrieval corpus consists of 84 long financial documents averaging {$\sim$}116K tokens each ({$\sim$}140 pages per PDF). Our agentic approach with GPT-5-mini achieves \textbf{92.00\%} correctness, substantially outperforming both traditional RAG (by \textbf{3.8$\times$}) and moreover, is more general than the agentic tool use baseline of \cite{Subramanian2026} (by \textbf{2.8$\times$}), which relies on keyword search tools like pdfgrep, rga, and Linux commands. We also adopt a baseline where the ground-truth full-page evidence is provided directly to GPT-5-mini, bypassing agentic retrieval entirely. This oracle setting achieves 94.00\% and establishes an upper bound on the model's reasoning ability given perfect evidence. Our agentic system is within 2 pp of this upper bound which demonstrates its effectiveness. Between GPT-5-mini vs Claude Sonnet 4.5, our harness is equally effective. 

\subsection{Token Cost and Retrieval Efficiency}
\label{sec:results_token_cost}

\begin{table*}[htbp]
\centering
\small
\caption{Total token usage comparison between AgenticRAG and Single-shot Search across BRIGHT splits and FinanceBench. All values are averages per query (in thousands) and include system prompt, tool calls, tool results, and any thinking tokens. Cost ratio is AgenticRAG total tokens divided by Single-shot Search total tokens.}
\label{tab:token_usage}
\begin{tabular}{lrrrr}
\toprule
 & & \multicolumn{2}{c}{Avg. Total Tokens (K)} & Cost \\
\cmidrule(lr){3-4}
Dataset & N & AgenticRAG & Single-shot & Ratio \\
\midrule
Biology & 103 & 49.2 & 23.0 & 2.1$\times$ \\
Earth Science & 116 & 52.9 & 18.7 & 2.8$\times$ \\
Economics & 103 & 44.6 & 21.3 & 2.1$\times$ \\
Psychology & 101 & 58.2 & 24.8 & 2.3$\times$ \\
Robotics & 101 & 56.2 & 23.1 & 2.4$\times$ \\
Stack Overflow & 117 & 55.7 & 15.8 & 3.5$\times$ \\
Sustainable Living & 108 & 59.3 & 19.9 & 3.0$\times$ \\
Pony & 112 & 42.6 & 17.6 & 2.4$\times$ \\
\midrule
\textbf{BRIGHT Avg.} & \textbf{861} & \textbf{52.3} & \textbf{20.4} & \textbf{2.6$\times$} \\
\midrule
FinanceBench & 150 & 114.8 & 14.7 & 7.8$\times$ \\
\bottomrule
\end{tabular}
\end{table*}

Table~\ref{tab:token_usage} quantifies the end-to-end token cost of agentic retrieval. We measure total tokens consumed across the full interaction, including model thinking, tool-call arguments, retrieved tool results, and final answer generation. On BRIGHT, AgenticRAG averages 52.3K total tokens per query, compared to 20.4K for Single-shot Search, a 2.6$\times$ token overhead. This cost yields a disproportionate quality gain: Claude Sonnet 4.5 with AgenticRAG reaches 49.6\% recall@1, compared to 8.41\% for Single-shot Search, a 5.9$\times$ improvement. FinanceBench is more expensive, averaging 114.8K tokens per query and a 7.8$\times$ ratio over single-shot search, which reflects deep navigation over long financial filings. This higher cost is paired with 92.00\% answer correctness, close to the 94.00\% oracle evidence upper bound. Tool usage in Table~\ref{tab:ablation_agentic} further shows that the system operates well within the 15-iteration budget, averaging 4.48--4.79 tool calls per query. The multi-query ablation provides a direct efficiency comparison: the full system achieves comparable recall with 4.79 average tool calls versus 6.79 without multi-query search, a 29\% reduction in tool calls.

\subsection{Ablation Studies}
\label{sec:results_ablation}

\begin{table*}[htbp]
\centering
\small
\caption{Ablation study of agentic components averaged across all BRIGHT splits. Performance is measured by recall (R@k), along with average tool usage and per-tool statistics. Per-split results are in Table~\ref{tab:ablation_agentic_full}.}
\label{tab:ablation_agentic}
\begin{tabular}{lccccccc}
\toprule
Variant & R@1 & R@3 & Avg. Tools & Search & Open & Find & Summ. \\
\midrule
Single-shot Search & 8.41$\pm$4.83 & 12.90$\pm$5.87 & 1 & 1 & - & - & - \\
Claude Sonnet 4.5 & 49.59$\pm$7.79 & 64.20$\pm$7.13 & 4.48$\pm$0.26 & 2.51$\pm$0.19 & 1.54$\pm$0.15 & 0.42$\pm$0.11 & 0.01$\pm$0.02 \\
GPT-5-mini & 43.49$\pm$8.00 & 62.53$\pm$7.23 & 4.79$\pm$0.71 & 3.39$\pm$0.55 & 1.22$\pm$0.25 & 0.14$\pm$0.08 & 0.06$\pm$0.05 \\
$\llcorner$ w/o Summarize & 43.34$\pm$8.07 & 63.85$\pm$7.21 & 4.92$\pm$0.70 & 3.44$\pm$0.54 & 1.31$\pm$0.27 & 0.16$\pm$0.09 & - \\
$\llcorner$ w/o Semantic Find & 46.34$\pm$7.97 & 64.44$\pm$7.07 & 5.02$\pm$0.73 & 3.47$\pm$0.55 & 1.34$\pm$0.27 & 0.17$\pm$0.09 & 0.06$\pm$0.05 \\
$\llcorner$ w/o Multi-query Search & 44.84$\pm$8.08 & 62.30$\pm$7.26 & 6.79$\pm$0.70 & 4.38$\pm$0.57 & 2.16$\pm$0.27 & 0.24$\pm$0.11 & 0.03$\pm$0.04 \\
\bottomrule
\end{tabular}
\end{table*}
\paragraph{Single-shot vs Agentic Retrieval.}
From Table~\ref{tab:ablation_agentic} the most significant finding is the dramatic improvement from single-shot search to full agentic tool use. Single-shot search achieves only 8.41\% recall@1 on average, while agentic tool use reaches \textbf{43.49\%} with GPT-5-mini and \textbf{49.59\%} with Claude Sonnet 4.5---representing \textbf{5.2$\times$} and \textbf{5.9$\times$} improvements respectively. Notably, the proprietary search stack behind our search tool trades off raw retrieval quality for speed, immense scale, and availability compared to the state-of-the-art embedding-based retrievers in Table~\ref{tab:bright_model_compare_main}. However, these quality differences vanish when our agentic harness is employed with a reasoning language model. The improvement is consistent across splits (ref. Table~\ref{tab:ablation_agentic_full}).

\paragraph{Model Comparison and Tool Usage Patterns.}
Claude Sonnet 4.5 achieves a \textbf{+6.1 pp} improvement over GPT-5-mini, outperforming on seven of eight splits (detailed per-split results in Appendix Table~\ref{tab:ablation_agentic_full}). The two models exhibit distinct strategies that reflect an exploration--exploitation trade-off. Claude favors \emph{exploitation}: it uses fewer search calls (2.51 vs 3.39) but opens more documents (1.54 vs 1.22) and relies more on semantic find (0.42 vs 0.14, a 3$\times$ increase), going deeper into candidate documents. GPT-5-mini favors \emph{exploration}: it issues more search calls with reformulated queries rather than using in-document find, casting a wider net across the corpus. In the BRIGHT long-document setting, where queries have only {$\sim$}1.9 golden documents on average amid a large corpus, relevant documents are sparse and broad exploration often surfaces irrelevant results.

\paragraph{Failure Patterns.}
The main observed weakness is broad multi-evidence retrieval, especially the Pony split, where each query has {$\sim$}6.9 gold documents on average compared to {$\sim$}1.9 across BRIGHT overall. This setting rewards recovering many related documents, whereas our harness is optimized for coarse-to-fine navigation toward a small number of high-value evidence sources. This explains why Pony remains difficult for both of our models despite large gains on scientific and technical splits where relevant documents are sparse. The pattern suggests that future trajectory policies should better detect broad evidence needs and shift from depth-first document reading to wider coverage before final ranking.

\paragraph{Component Contributions.}
We ablate individual components using GPT-5-mini to understand their contributions. The most notable finding concerns multi-query search. In the default configuration, the model can issue up to 5 queries in parallel within a single search tool call, with results de-duped and presented together. Restricting the system to single-query search (\textbf{w/o Multi-query Search}, 44.84\%), where the model issues only one query per search call but receives the same number of results, achieves comparable recall@1 to the full system but at the cost of increased tool usage---6.79 average tool calls compared to 4.79 for the full system, with notably more search operations (4.38 vs 3.39) and document opens (2.16 vs 1.22). This suggests that multi-query search improves efficiency by finding relevant documents with fewer iterations. Detailed analysis of these ablations is provided in Appendix~\ref{app:detailed_ablation}.

\paragraph{Findings from Pre-Production Deployments}
In our pre-production evaluation we identified several design choices that guide the model toward more optimal trajectories: (1)~\textbf{Surfacing document metadata in search results} like title, filename, and file type helps the model disambiguate semantically similar snippets and avoid redundant searches; (2)~\textbf{Line-numbered document previews} lets the model anchor on specific content and jump to relevant sections in successive open calls; (3)~\textbf{Candidate reference retention after summarization} in the context window enables the model to go deeper on promising candidates via open/find, rather than restarting retrieval with reformulated queries.(4)~\textbf{Having a switcher} route complex, multi-intent queries to our agentic rag harness for deeper analysis. Simple queries are routed to traditional rag for faster answers. This is vital for the tradeoff between user experience, cost, model availability. We have good early signals of this being effective and we continue to pursue this hybrid approach.

\section{Conclusion}
\label{sec:conclusion}
    We presented a practical harness for AgenticRAG that equips reasoning language models with search, find, open, and summarize tools to autonomously retrieve and reason over large enterprise corpora. Across three benchmarks, our approach achieves 49.6\% recall@1 on BRIGHT (+21.8 pp over the best embedding baseline), 0.96 factuality on WixQA (+13\% relative), and 92.00\% answer correctness on FinanceBench---within 2 pp of oracle access. Token analysis shows that these gains require a moderate 2.6$\times$ token overhead on BRIGHT relative to single-shot search, while delivering a 5.9$\times$ recall@1 improvement. These results demonstrate that our harness effectively extracts the value of reasoning models for enterprise information retrieval tasks requiring deep, multi-step reasoning. Future work will focus on large-scale deployment, budget-aware routing between traditional and agentic RAG, deeper failure analysis, ablations over iteration and window-size budgets, and optimizing retrieval trajectories for fast iterative reasoning via fine tuning.

\section*{Acknowledgments}
We thank Eli Coon, Kinfe Mengistu and members of the broader Copilot Studio team for feedback and discussions during internal experimentations. Special thanks to James Cai and Alejandro Gutierrez Munoz for technical guidance and project sponsorship.

\bibliography{main-bib}

\appendix

\section{Method Details}
\label{app:method}

\subsection{Agentic Loop Algorithm}
\label{app:agentic_loop_algo}

Detailed agentic loop algorithm is shown in Algorithm \ref{algo:agentic_loop}.

\begin{algorithm}
\small
\caption{Agentic Loop}
\label{algo:agentic_loop}
\begin{algorithmic}[1] 
\Require $\text{user\_query}$, $\text{max\_calls}$, $\text{token\_threshold}$
\Ensure Formatted answer with citations

\State $\text{conversation}.\text{add}(\text{user\_query})$
\For{$i = 1$ \textbf{to} $\text{max\_calls}$}
    \If{$\text{tokens}(\text{conversation}) \geq \text{token\_threshold}$}
        \State \Call{ManageContext}{ } \Comment{Force summarize}
    \EndIf
    \State $\text{response} \gets \text{LLM}(\text{conversation}, \text{tool\_schemas})$
    \If{$\text{response}.\text{has\_tool\_calls}$}
        \For{\textbf{each} $\text{tool\_call}$ \textbf{in} $\text{response}.\text{tool\_calls}$}
            \State $\text{result} \gets \Call{ExecuteTool}{\text{tool\_call}}$
            \State $\text{conversation}.\text{add}(\text{tool\_call}, \text{result})$
        \EndFor
    \Else
        \State \Return $\Call{FormatAnswer}{\text{response}.\text{text}}$
    \EndIf
\EndFor
\State \Return \Call{ForceFinalAnswer}{ }
\end{algorithmic}
\end{algorithm}

\subsection{System Instructions for Tool Use}
\label{app:system_prompt}

Overall instructions include: 
\begin{itemize}[noitemsep]
    \item Search before answering when uncertain.
    \item Progressively explore using find or open when snippets are insufficient.
    \item Reuse previous results rather than performing search again.
    \item Cite every time when information is used from tool outputs.
\end{itemize}
When to use search: 
\begin{itemize}[noitemsep]
    \item Primary search tool across enterprise corpus.
    \item First choice for any work-related query.
    \item When users reference current/changing information, enterprise-specific terms, or acronyms.
    \item To verify details rather than making assumptions.
\end{itemize}
When to use find:
\begin{itemize}[noitemsep]
    \item In-document pattern search for relevant files from search results.
    \item When search results do not give enough details.
    \item To get a focused view of a result in relation to certain terms.
\end{itemize}
When to use open:
\begin{itemize}[noitemsep]
    \item Windowed full content retrieval for relevant files from search results.
    \item When search results snippets are insufficient.
    \item To pull in more content from the most promising results.
    \item Can open multiple search results.
    \item Option to choose a line number close to the relevant content.
\end{itemize}

\begin{table*}[t]
\centering
\small
\caption{Dataset statistics for the BRIGHT benchmark \cite{su2024bright} long-context splits used in our evaluation.}
\label{tab:bright_dataset_stats}
\begin{tabular}{lccccc}
\toprule
Split & \# Query & \# Docs & Avg Query Len & Avg Doc Len & Avg \# Gold Docs \\
\midrule
Biology            & 103 &   524 &  115.2 &  9,422.4 & 1.3 \\
Earth Science      & 116 &   601 &  109.5 & 27,312.3 & 1.6 \\
Economics          & 103 &   516 &  181.5 & 11,896.4 & 1.1 \\
Psychology         & 101 &   512 &  149.6 & 12,411.7 & 1.1 \\
Robotics           & 101 &   508 &  818.9 & 14,998.2 & 1.1 \\
Stack Overflow     & 117 & 1,858 &  478.3 & 40,759.7 & 1.1 \\
Sustainable Living & 108 &   554 &  148.5 & 12,077.7 & 1.2 \\
Pony               & 112 &   577 &  102.6 &  1,361.0 & 6.9 \\
\midrule
\textbf{Total/Avg} & 861 & 5,650 &  263.0 & 16,279.9 & 1.9 \\
\bottomrule
\end{tabular}
\end{table*}

\begin{table*}[t]
\centering
\small
\caption{Dataset statistics for the WixQA \cite{cohen2025wixqa} benchmark (median values). }
\label{tab:wixqa_dataset_stats}
\begin{tabular}{lcccc}
\toprule
Dataset & \# Query & \# Query Tokens & \# Answer Tokens & Multi-Article \% \\
\midrule
ExpertWritten & 200 & 19 & 172 & 27\% \\
Simulated     & 200 & 12 &  50 & 14\% \\
\bottomrule
\end{tabular}
\end{table*}

\section{Dataset Details}
\label{app:datasets}

\subsection{BRIGHT Benchmark}
\label{app:bright_details}
We adopt the BRIGHT benchmark \cite{su2024bright}, which is designed to capture realistic enterprise scenarios of information retrieval. BRIGHT derives queries from StackExchange posts, reflecting human-authored, highly situational and domain-specific information needs. For each query, the corpus contains positive documents cited in top-voted answers and verified by human annotators, as well as negative documents collected via search engine retrieval. The corpora is normalized web content (e.g., Wikipedia pages, blogs, and reports). This construction has shown to yields realistic retrieval pools with substantial semantic overlap between relevant and irrelevant documents. We choose to evaluate on the long-context setting of BRIGHT, where documents correspond to entire web pages rather than snippets and the task is to retrieve the full relevant document(s) for a given query. Our experiments span eight domains: Biology, Earth Science, Economics, Psychology, Robotics, Stack Overflow, Sustainable Living, and Pony. These domains cover a broad range of scientific, technical, and professional areas commonly encountered in enterprise information retrieval. Across domains, corpora contain hundreds to thousands of documents, with average document lengths ranging from several thousand to over 40k tokens. Queries themselves are also non-trivial in length, with average query sizes exceeding 100 tokens in most domains and reaching several hundred tokens in technical domains such as Robotics and Stack Overflow. Benchmark statistics are detailed in Table \ref{tab:bright_dataset_stats}. We follow the standard evaluation protocol of the BRIGHT benchmark and report Recall@1 for long-context document retrieval.

\subsection{WixQA Benchmark}
\label{app:wixqa_details}
WixQA targets procedural, long-form queries that require multi-step reasoning and specialized enterprise vocabulary, closely matching real-world support and troubleshooting scenarios. We utilize both the subsets in WiXQA for our experiments: \textit{Expert Written}, containing authentic customer queries with step-by-step answers authored and validated by human domain experts, and \textit{Simulated}, derived from multi-turn user–chatbot interactions and curated into single-turn queries with expert-validated procedural correctness. A defining characteristic of WixQA is its multi-article dependency, where answering a query may require retrieving and synthesizing information from multiple documents. All queries are grounded in a shared enterprise-scale knowledge base of 6,221 domain-specific help articles, making WixQA well suited for evaluating agentic RAG that must coordinate retrieval and reasoning over complex, multi-document enterprise corpora. Datasets statistics are presented in \ref{tab:wixqa_dataset_stats}

\subsection{FinanceBench}
\label{app:financebench_details}
\begin{table}[htbp]
\centering
\small
\caption{FinanceBench ~\cite{islam2023financebench} data statistics.}
\label{tab:financebench_stats}
\begin{tabular}{lc}
\toprule
Statistic & Value \\
\midrule
\# Queries & 150 \\
\# Ground docs & 84 \\
Avg. \# pages / doc & 143 \\
Avg. \# tokens / doc & 116,715 \\
Total tokens & 9,804,065 \\
\bottomrule
\end{tabular}
\end{table}
FinanceBench \cite{islam2023financebench} is a human-evaluated benchmark consisting of financial questions over public company filings (10-K, 10-Q, 8-K, and earnings reports in PDF form). Questions span metrics-generated and domain-relevant categories: metrics-generated questions target specific financial line items or ratios that require the model to locate the relevant data in the document and often perform a calculation, making them straightforward to verify since each has a single unambiguous answer. Domain-relevant questions require deeper financial reasoning, such as identifying drivers of margin changes or assessing capital intensity. Each query pertains to a single document. We choose this benchmark because of the large size of its documents (averaging {$\sim$}143 pages and {$\sim$}117K tokens per PDF), which is representative of enterprise domains where knowledge workers routinely work with dense, information-heavy manuals, reports, and regulatory filings. Corpus statistics are presented in Table~\ref{tab:financebench_stats}. The evaluation metric we use is answer correctness. LLM as a judge is used for this process and manual review of the results is also conducted.

\section{Additional Results}

\subsection{Full BRIGHT Retrieval Results}
\label{app:bright_full_results}

\begin{table*}[htbp]
\centering
\small
\caption{Full long-context retrieval performance on unsplit web pages of StackExchange data from BRIGHT benchmark. Scores are reported in recall@1.}
\label{tab:bright_model_compare_full}
\begin{tabular}{lccccccccc}
\toprule
 & Bio. & Earth. & Econ. & Psy. & Rob. & Stack. & Sus. & Pony & Avg. \\
\midrule
\multicolumn{10}{c}{\textbf{Sparse models}} \\
\midrule
BM25 & 10.7 & 15.4 & 10.7 & 8.4 & 7.4 & 22.2 & 10.7 & 5.4 & 11.4 \\
\midrule
\multicolumn{10}{c}{\textbf{Open-sourced embedding models}} \\
\midrule
BGE & 16.4 & 27.7 & 20.9 & 11.6 & 10.9 & 13.3 & 16.9 & 0.4 & 14.8 \\
Inst-L & 24.6 & 29.9 & 13.1 & 20.3 & 12.9 & 15.0 & 25.4 & 3.9 & 18.1 \\
SBERT & 25.6 & 34.1 & 18.9 & 15.8 & 10.9 & 15.0 & 18.0 & 1.2 & 17.4 \\
E5 & 29.9 & 36.3 & 26.2 & 46.7 & 17.3 & 14.5 & 32.2 & 1.1 & 25.5 \\
SFR & 30.3 & 37.0 & 24.3 & 47.7 & 17.3 & 14.5 & 35.0 & 2.0 & 26.0 \\
Inst-XL & 21.5 & 31.0 & 13.1 & 20.5 & 13.9 & 15.0 & 20.1 & 6.0 & 17.6 \\
GritLM & 37.5 & 40.3 & 25.7 & 34.4 & 17.8 & 20.1 & 32.4 & 0.0 & 26.0 \\
Qwen & 39.2 & 36.1 & 25.7 & 42.3 & 21.3 & 23.5 & 33.1 & 1.3 & 27.8 \\
\midrule
\multicolumn{10}{c}{\textbf{Proprietary embedding models}} \\
\midrule
Cohere & 31.5 & 34.5 & 18.9 & 20.5 & 9.9 & 15.8 & 15.2 & 0.8 & 18.4 \\
OpenAI & 32.1 & 31.4 & 23.8 & 34.2 & 11.9 & 10.7 & 26.3 & 0.0 & 21.3 \\
Voyage & 34.4 & 35.4 & 26.7 & 41.6 & 12.9 & 12.8 & 31.1 & 1.3 & 24.5 \\
Google & 30.9 & 38.0 & 21.9 & 30.7 & 12.9 & 19.2 & 25.7 & 0.3 & 22.4 \\
\midrule
\multicolumn{10}{c}{\textbf{Reasoning enhanced methods}} \\
\midrule
Claude-3-Opus (BM25) & 26.8 & 13.5 & 13.4 & 28.2 & 7.9 & 28.2 & 11.8 & -- & 18.5 \\
GPT-4 (BM25) & 26.8 & 15.8 & 10.2 & 30.7 & 5.9 & 26.5 & 9.7 & -- & 17.9 \\
DeepSeek-R1 (BM25) & 26.8 & 20.0 & 14.4 & 30.2 & 14.9 & 33.3 & 10.6 & -- & 21.5 \\
ReDI (BM25) & 28.4 & 22.4 & 21.2 & 32.0 & 19.8 & 36.3 & 21.7 & -- & 26.0 \\
Claude-3-Opus (SBERT) & 34.8 & 31.6 & 21.8 & 15.8 & 8.9 & 15.8 & 16.6 & -- & 20.8 \\
GPT-4 (SBERT) & 37.7 & 35.3 & 19.9 & 18.3 & 12.4 & 11.5 & 22.6 & -- & 22.5 \\
DeepSeek-R1 (SBERT) & 35.6 & 34.8 & 16.0 & 15.3 & 8.9 & 15.0 & 19.9 & -- & 20.8 \\
ReDI (SBERT) & 36.2 & 32.8 & 22.8 & 20.8 & 10.9 & 16.2 & 22.2 & -- & 23.1 \\
\midrule
\multicolumn{10}{c}{\textbf{Our Agentic methods (search/find/open)}} \\
\midrule
GPT-5-mini & 61.7 & 48.1 & 41.4 & 65.3 & 39.4 & \textbf{40.6} & 46.6 & 4.8 & 43.5 \\
\textbf{Claude Sonnet 4.5} & \textbf{62.3} & \textbf{60.0} & \textbf{58.7} & \textbf{67.9} & \textbf{55.0} & 34.1 & \textbf{51.7} & \textbf{7.1} & \textbf{49.6} \\
\bottomrule
\end{tabular}
\end{table*}

Table~\ref{tab:bright_model_compare_full} presents the complete retrieval results on the BRIGHT benchmark across all baseline models.

\subsection{Detailed Ablation Analysis}
\label{app:detailed_ablation}

\begin{table*}[htbp]
\centering
\small
\caption{Per-split ablation study of agentic components on BRIGHT. Performance is measured by recall (R@k), along with average tool usage and per-tool usage statistics.}
\label{tab:ablation_agentic_full}
\begin{tabular}{lccccccc}
\toprule
Variant & R@1 & R@3 & Avg. Tools & Search & Open & Find & Sum. \\
\midrule
\multicolumn{8}{c}{\textbf{Bio.}} \\
\midrule
Single-shot Search & 10.88$\pm$5.70 & 14.63$\pm$6.55 & 1 & 1 & - & - & - \\
Claude Sonnet 4.5 & \textbf{62.34$\pm$8.33} & 80.47$\pm$7.19 & 4.36$\pm$0.28 & 2.12$\pm$0.20 & 1.72$\pm$0.14 & 0.52$\pm$0.12 & 0.01$\pm$0.02 \\
GPT-5-mini & 61.72$\pm$8.66 & 88.28$\pm$5.53 & 4.54$\pm$0.78 & 3.46$\pm$0.60 & 0.98$\pm$0.25 & 0.09$\pm$0.06 & 0.02$\pm$0.03 \\
$\llcorner$ w/o Summarize & 58.16$\pm$8.59 & 83.33$\pm$6.68 & 4.52$\pm$0.80 & 3.38$\pm$0.57 & 1.06$\pm$0.30 & 0.07$\pm$0.05 & - \\
$\llcorner$ w/o Semantic Find & 59.43$\pm$8.42 & 86.53$\pm$6.31 & 4.49$\pm$0.71 & 3.32$\pm$0.57 & 1.11$\pm$0.28 & 0.04$\pm$0.04 & 0.02$\pm$0.03 \\
$\llcorner$ w/o Multi-query & 60.35$\pm$8.86 & 85.44$\pm$6.40 & 6.63$\pm$0.80 & 4.34$\pm$0.64 & 2.20$\pm$0.30 & 0.09$\pm$0.07 & - \\
\midrule
\multicolumn{8}{c}{\textbf{Earth.}} \\
\midrule
Single-shot Search & 7.35$\pm$4.48 & 13.62$\pm$6.09 & 1 & 1 & - & - & - \\
Claude Sonnet 4.5 & \textbf{60.01$\pm$7.38} & 78.51$\pm$5.99 & 4.54$\pm$0.25 & 2.28$\pm$0.19 & 1.70$\pm$0.14 & 0.54$\pm$0.12 & 0.01$\pm$0.02 \\
GPT-5-mini & 48.10$\pm$8.02 & 72.06$\pm$7.30 & 5.06$\pm$0.78 & 3.52$\pm$0.60 & 1.36$\pm$0.27 & 0.12$\pm$0.07 & 0.07$\pm$0.05 \\
$\llcorner$ w/o Summarize & 48.44$\pm$8.33 & 72.66$\pm$6.93 & 4.88$\pm$0.69 & 3.36$\pm$0.55 & 1.41$\pm$0.27 & 0.11$\pm$0.07 & - \\
$\llcorner$ w/o Semantic Find & 56.10$\pm$7.87 & 75.93$\pm$6.25 & 4.99$\pm$0.74 & 3.41$\pm$0.56 & 1.38$\pm$0.28 & 0.14$\pm$0.09 & 0.06$\pm$0.04 \\
$\llcorner$ w/o Multi-query & 54.43$\pm$7.73 & 75.00$\pm$6.61 & 6.81$\pm$0.71 & 4.30$\pm$0.58 & 2.28$\pm$0.27 & 0.20$\pm$0.10 & 0.03$\pm$0.04 \\
\midrule
\multicolumn{8}{c}{\textbf{Econ.}} \\
\midrule
Single-shot Search & 17.39$\pm$7.61 & 21.74$\pm$8.15 & 1 & 1 & - & - & - \\
Claude Sonnet 4.5 & \textbf{58.74$\pm$9.47} & 68.28$\pm$8.74 & 4.50$\pm$0.24 & 2.37$\pm$0.15 & 1.62$\pm$0.16 & 0.50$\pm$0.13 & 0.01$\pm$0.02 \\
GPT-5-mini & 41.41$\pm$9.60 & 67.68$\pm$9.09 & 3.65$\pm$0.69 & 2.52$\pm$0.51 & 1.04$\pm$0.27 & 0.06$\pm$0.06 & 0.03$\pm$0.03 \\
$\llcorner$ w/o Summarize & 45.96$\pm$9.85 & 72.22$\pm$8.84 & 4.07$\pm$0.79 & 3.05$\pm$0.63 & 0.90$\pm$0.26 & 0.12$\pm$0.10 & - \\
$\llcorner$ w/o Semantic Find & 46.00$\pm$9.50 & 67.50$\pm$9.25 & 3.97$\pm$0.77 & 2.99$\pm$0.58 & 0.86$\pm$0.25 & 0.05$\pm$0.05 & 0.07$\pm$0.07 \\
$\llcorner$ w/o Multi-query & 42.27$\pm$9.54 & 67.53$\pm$9.28 & 6.32$\pm$0.70 & 4.00$\pm$0.59 & 2.20$\pm$0.29 & 0.12$\pm$0.07 & - \\
\midrule
\multicolumn{8}{c}{\textbf{Psy.}} \\
\midrule
Single-shot Search & 4.88$\pm$4.27 & 6.95$\pm$5.24 & 1 & 1 & - & - & - \\
Claude Sonnet 4.5 & \textbf{67.86$\pm$8.93} & 83.57$\pm$6.99 & 4.08$\pm$0.24 & 2.08$\pm$0.17 & 1.62$\pm$0.15 & 0.37$\pm$0.11 & 0.01$\pm$0.02 \\
GPT-5-mini & 65.26$\pm$8.95 & 78.84$\pm$7.89 & 3.87$\pm$0.68 & 2.82$\pm$0.54 & 0.86$\pm$0.23 & 0.07$\pm$0.06 & 0.12$\pm$0.07 \\
$\llcorner$ w/o Summarize & 58.95$\pm$9.47 & 81.26$\pm$7.63 & 3.97$\pm$0.65 & 2.89$\pm$0.50 & 1.03$\pm$0.27 & 0.04$\pm$0.04 & - \\
$\llcorner$ w/o Semantic Find & 65.26$\pm$9.21 & 83.89$\pm$6.89 & 4.36$\pm$0.76 & 2.89$\pm$0.53 & 1.29$\pm$0.31 & 0.05$\pm$0.05 & 0.12$\pm$0.07 \\
$\llcorner$ w/o Multi-query & 67.06$\pm$9.42 & 80.04$\pm$8.04 & 6.51$\pm$0.77 & 4.08$\pm$0.65 & 2.27$\pm$0.33 & 0.13$\pm$0.08 & 0.02$\pm$0.03 \\
\midrule
\multicolumn{8}{c}{\textbf{Rob.}} \\
\midrule
Single-shot Search & 10.20$\pm$6.12 & 15.31$\pm$7.14 & 1 & 1 & - & - & - \\
Claude Sonnet 4.5 & \textbf{54.95$\pm$9.41} & 71.29$\pm$8.66 & 4.92$\pm$0.33 & 2.97$\pm$0.22 & 1.54$\pm$0.16 & 0.41$\pm$0.12 & - \\
GPT-5-mini & 39.39$\pm$9.34 & 60.61$\pm$9.34 & 5.22$\pm$0.75 & 3.65$\pm$0.56 & 1.44$\pm$0.29 & 0.13$\pm$0.09 & - \\
$\llcorner$ w/o Summarize & 39.80$\pm$9.18 & 63.27$\pm$9.44 & 5.13$\pm$0.69 & 3.54$\pm$0.56 & 1.46$\pm$0.29 & 0.13$\pm$0.07 & - \\
$\llcorner$ w/o Semantic Find & 43.68$\pm$9.74 & 65.26$\pm$9.21 & 5.63$\pm$0.78 & 3.78$\pm$0.59 & 1.61$\pm$0.30 & 0.22$\pm$0.12 & 0.02$\pm$0.03 \\
$\llcorner$ w/o Multi-query & 46.94$\pm$9.69 & 65.82$\pm$9.44 & 6.95$\pm$0.68 & 4.62$\pm$0.58 & 2.09$\pm$0.25 & 0.24$\pm$0.10 & - \\
\midrule
\multicolumn{8}{c}{\textbf{Stack.}} \\
\midrule
Single-shot Search & 7.28$\pm$4.61 & 12.14$\pm$6.07 & 1 & 1 & - & - & - \\
Claude Sonnet 4.5 & 34.05$\pm$8.19 & 43.53$\pm$8.62 & 4.76$\pm$0.28 & 3.31$\pm$0.26 & 1.07$\pm$0.15 & 0.36$\pm$0.11 & 0.02$\pm$0.02 \\
GPT-5-mini & 40.62$\pm$8.71 & 53.12$\pm$8.93 & 6.03$\pm$0.76 & 4.88$\pm$0.65 & 0.85$\pm$0.18 & 0.25$\pm$0.15 & 0.05$\pm$0.05 \\
$\llcorner$ w/o Summarize & 37.84$\pm$8.56 & 54.95$\pm$9.01 & 6.71$\pm$0.75 & 5.11$\pm$0.64 & 1.26$\pm$0.21 & 0.34$\pm$0.15 & - \\
$\llcorner$ w/o Semantic Find & \textbf{41.52$\pm$8.71} & 54.02$\pm$8.93 & 6.62$\pm$0.78 & 5.20$\pm$0.66 & 1.18$\pm$0.21 & 0.19$\pm$0.10 & 0.05$\pm$0.04 \\
$\llcorner$ w/o Multi-query & 33.77$\pm$8.55 & 43.42$\pm$8.77 & 7.93$\pm$0.71 & 6.25$\pm$0.63 & 1.24$\pm$0.19 & 0.43$\pm$0.21 & - \\
\midrule
\multicolumn{8}{c}{\textbf{Sus.}} \\
\midrule
Single-shot Search & 8.87$\pm$5.51 & 18.28$\pm$7.26 & 1 & 1 & - & - & - \\
Claude Sonnet 4.5 & 51.65$\pm$8.97 & 69.15$\pm$8.19 & 4.48$\pm$0.21 & 2.32$\pm$0.15 & 1.72$\pm$0.14 & 0.44$\pm$0.11 & - \\
GPT-5-mini & 46.65$\pm$9.19 & 73.45$\pm$8.13 & 5.01$\pm$0.65 & 2.98$\pm$0.45 & 1.88$\pm$0.33 & 0.09$\pm$0.06 & 0.06$\pm$0.05 \\
$\llcorner$ w/o Summarize & 53.25$\pm$9.22 & 76.91$\pm$7.54 & 5.17$\pm$0.71 & 3.13$\pm$0.47 & 1.91$\pm$0.34 & 0.13$\pm$0.10 & - \\
$\llcorner$ w/o Semantic Find & \textbf{55.53$\pm$9.16} & 75.83$\pm$7.96 & 5.39$\pm$0.67 & 3.14$\pm$0.45 & 2.02$\pm$0.33 & 0.15$\pm$0.09 & 0.08$\pm$0.06 \\
$\llcorner$ w/o Multi-query & 50.53$\pm$9.54 & 76.47$\pm$8.02 & 7.25$\pm$0.61 & 4.06$\pm$0.51 & 2.92$\pm$0.27 & 0.23$\pm$0.10 & 0.04$\pm$0.05 \\
\midrule
\multicolumn{8}{c}{\textbf{Pony}} \\
\midrule
Single-shot Search & 0.40$\pm$0.35 & 0.52$\pm$0.46 & 1 & 1 & - & - & - \\
Claude Sonnet 4.5 & \textbf{7.12$\pm$1.63} & 18.79$\pm$2.67 & 4.22$\pm$0.29 & 2.62$\pm$0.19 & 1.36$\pm$0.12 & 0.25$\pm$0.10 & - \\
GPT-5-mini & 4.79$\pm$1.54 & 6.20$\pm$1.63 & 4.93$\pm$0.58 & 3.31$\pm$0.45 & 1.35$\pm$0.22 & 0.27$\pm$0.11 & - \\
$\llcorner$ w/o Summarize & 4.34$\pm$1.38 & 6.16$\pm$1.63 & 4.89$\pm$0.56 & 3.09$\pm$0.42 & 1.45$\pm$0.21 & 0.35$\pm$0.18 & - \\
$\llcorner$ w/o Semantic Find & 3.19$\pm$1.18 & 6.58$\pm$1.76 & 4.72$\pm$0.63 & 3.00$\pm$0.44 & 1.22$\pm$0.22 & 0.50$\pm$0.17 & - \\
$\llcorner$ w/o Multi-query & 3.34$\pm$1.27 & 4.66$\pm$1.51 & 5.92$\pm$0.63 & 3.35$\pm$0.41 & 2.07$\pm$0.23 & 0.50$\pm$0.17 & - \\
\bottomrule
\end{tabular}
\end{table*}

Table~\ref{tab:ablation_agentic_full} provides per-split ablation results across all BRIGHT domains. Removing the summarization tool (\textbf{w/o Summarize}, 43.34\% avg recall@1) has minimal impact, indicating that this component is rarely needed for the retrieval task. Removing semantic find (\textbf{w/o Semantic Find}, 46.34\%) slightly \emph{improves} average recall@1, likely because the lexical find fallback is sufficient for most in-document searches and removing the semantic option reduces latency, allowing more search iterations within the same compute budget.

\subsection{WixQA Simulated Results}
\label{app:wixqa_simulated_results}

\begin{figure}[]
\centering
\includegraphics[width=0.8\columnwidth]{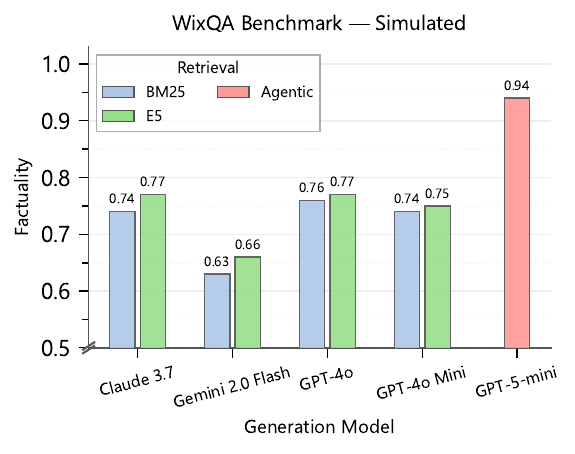}
\caption{Factuality performance on the WixQA Simulated dataset. The performance gap between agentic retrieval and traditional methods is even larger on synthetic questions that require more complex reasoning.}
\label{fig:wixqa_simulated}
\end{figure}

As shown in Figure. \ref{fig:wixqa_simulated}, on simulated questions with expert validated ground truth answers i.e. \textit{Simulated} split of WixQA, our method achieves \textbf{0.94} factuality, compared to 0.77 for both E5+GPT-4o and E5+Claude 3.7. The improvement is even more pronounced on this dataset, with a \textbf{22\%} relative gain. This suggests that agentic retrieval is particularly effective when questions require more complex reasoning or multi-hop information gathering.





\subsection{Example Conversation}

\begin{figure}[t]
\centering
\includegraphics[width=\columnwidth]{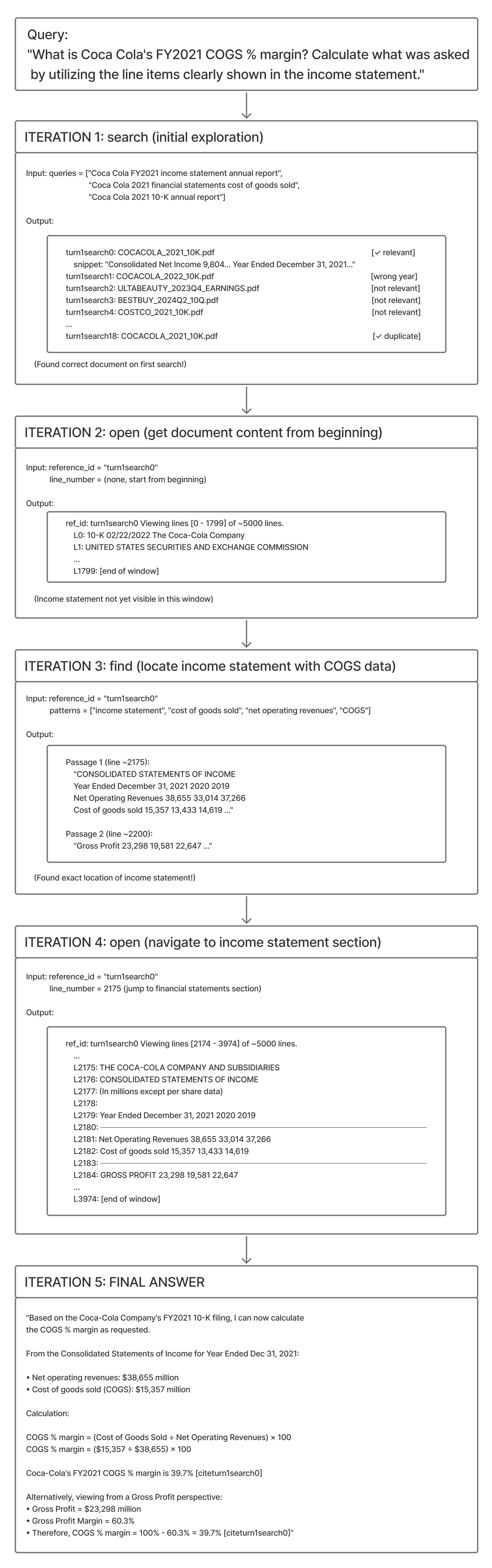}
\caption{Example conversation from FinanceBench.}
\label{fig:example_conversation}
\end{figure}

Figure \ref{fig:example_conversation} shows an example conversation from the FinanceBench. 

\end{document}